\pdfoutput=1
\documentclass[11pt]{article}

\usepackage{naacl2021}
\usepackage{amsthm}
\usepackage{latexsym}
\usepackage{graphicx}
\usepackage{subfigure} 
\usepackage{url}
\usepackage{enumerate}
\usepackage{amsmath}
\usepackage{mathtools}
\usepackage{footnote}
\usepackage{amssymb}
\usepackage{xcolor}
\usepackage[utf8]{inputenc}
\usepackage{ascii}
\usepackage{newunicodechar}

\usepackage[ruled,vlined,noend]{algorithm2e}
\usepackage{nicefrac}
\usepackage{xfrac}

\usepackage{tabularx, booktabs, enumitem}
\newcolumntype{Y}{>{\centering\arraybackslash}X}
\newcolumntype{L}{>{\arraybackslash}X}
\newcolumntype{R}{>{\raggedleft\arraybackslash}X}

\usepackage{multirow}
\usepackage{xspace}

\usepackage{booktabs} % For formal tables
\newcommand{\hide}[1]{} %hide

\newcommand{\st}{\textit{s.t.}}

\newcommand{\tightcaption}{\vspace{-.3em}}

\newcommand{\method}{\textit{CompoundGrow}\xspace}

\definecolor{gred}{RGB}{219,68,55}
\definecolor{gblue}{RGB}{66,133,244}
\definecolor{gyellow}{RGB}{244,180,0}
\definecolor{ggreen}{RGB}{15,157,88}
\definecolor{ggrey}{RGB}{115,115,115}

\newcommand{\tightmath}[1]{\everymath{\medmuskip=1.5mu minus 1.5mu\thickmuskip=2mu minus 2mu}$#1$\everymath{\medmuskip=2mu minus 2mu\thickmuskip=4mu minus 4mu}}

\newcommand{\myfootnote}[2]{\footnote{\footnotesize{#1} \href{#2}{\footnotesize{#2}}}}
\input{smile}

\usepackage{times}
\usepackage{latexsym}

\usepackage[T1]{fontenc}
\usepackage[utf8]{inputenc}

\usepackage{microtype}

\title{On the Transformer Growth for Progressive BERT Training}

\author{Xiaotao Gu\textsuperscript{\ACK\EOT}\Thanks{Work done while interning at Google. Corresponding Author: Hongkun Yu and Xiaotao Gu.}\;\; Liyuan Liu\textsuperscript{\ACK}\; Hongkun Yu\textsuperscript{\EOT}\; Jing Li\textsuperscript{\EOT}\; Chen Chen\textsuperscript{\EOT}\; Jiawei Han\textsuperscript{\ACK} \\
\EOT Google Research ~~ \texttt{\footnotesize \{hongkuny,jingli,chendouble\}@google.com} \\
\ACK University of Illinois at Urbana-Champaign ~~\texttt{\footnotesize \{xiaotao2,ll2,hanj\}@illinois.edu}\\
}

\begin{document}
\maketitle

\begin{abstract}
% Liu: V2
Due to the excessive cost of large-scale language model pre-training, considerable efforts have been made to train BERT progressively---start from an inferior but low-cost model and gradually grow the model to increase the computational complexity. 
Our objective is to advance the understanding of Transformer growth and discover principles that guide progressive training.
First, we find that similar to network architecture search, Transformer growth also favors compound scaling. 
Specifically, while existing methods only conduct network growth in a single dimension, we observe that it is beneficial to use compound growth operators and balance multiple dimensions (e.g., depth, width, and input length of the model). 
Moreover, we explore alternative growth operators in each dimension via controlled comparison to give operator selection practical guidance.
In light of our analyses, the proposed method \emph{\method} speeds up BERT pre-training by $73.6\%$ and $82.2\%$ for the base and large models respectively, while achieving comparable performances\myfootnote{Code will be released at:}{{https://github.com/google-research/google-research/tree/master/grow\_bert}}. 
\end{abstract}

\section{Introduction}

Thanks to the rapid increase of computing power, large-scale pre-training has been breaking the glass ceiling for natural language processing tasks~\citep{liu_empower_2018,peters2018deep,devlin2019bert,Liu2019RoBERTaAR,Brown2020LanguageMA}. 
However, with great power comes great challenges: the required excessive computational consumption significantly impedes the efficient iteration of both research exploration and industrial application. 
To lower the training cost, many attempts have been made to conduct \emph{progressive training}, which starts from training an inferior but low-cost model, and gradually increases its resource consumption~\citep{Gong2019EfficientTO,devlin2019bert}.
As elaborated in Section \ref{sec:related},
two components are typically needed for designing such progressive training algorithms---the growth scheduler and the growth operator~\citep{dong_lipgrow_2020}. 
The former controls when to conduct network growth, and the latter controls how to perform network growth. 
Here, our objectives are to better understand growth operators with a focus on Transformer models~\citep{Vaswani2017AttentionIA,Liu2020UnderstandingTD}, and specifically, to help design better progressive algorithms for BERT pre-training~\citep{devlin2019bert}. 
Specifically, we recognize the importance of using \emph{compound growth operators} in our study, which balance different model dimensions (e.g., number of layers, the hidden size, and the input sequence length).

Regarding previous efforts made on Transformer growth, they mainly focus on one single model dimension: either the length \citep{devlin2019bert} or the depth \citep{Gong2019EfficientTO}.
In this work, however, we find that \emph{compound effect} plays a vital role in growing a model to different capacities, just like its importance in deciding network architectures under specific budgets \citep{tan2019efficientnet}.
Here, we show that growing a Transformer from both dimensions leads to better performance with less training cost, which verifies our intuition and shows the potential of using compound growth operators in progressive BERT training. 

\begin{figure*}[t]
\centering
    \includegraphics[width=.9\textwidth]{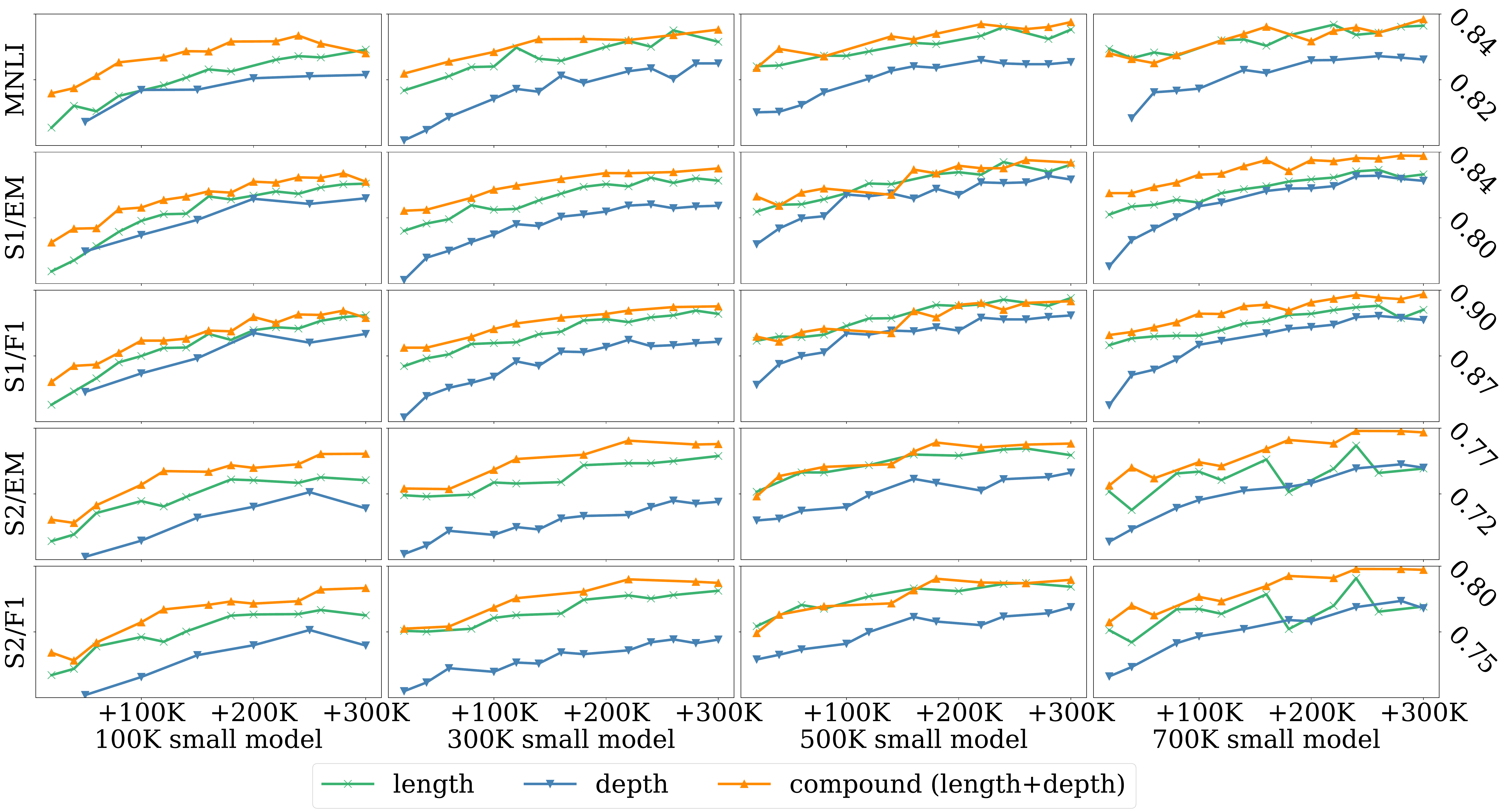}
\vspace{-0.2cm}
\caption{
Comparison of single-dimensional operators and the compound operator with comparable cost. 
Y-axis indicates finetuning performances, including MNLI-match valid accuracy (MNLI), SQuaD v1.1 exact match score and F1 (S1/EM, S1/F1), and SQuaD v2.0 exact match score and F1 (S2/EM, S2/F1).
X-axis stands for different training steps of the full model (12-layer BERT-base model with 512-token training data) in the last stage.
Different columns represent different training steps for the small (low-cost) model.
The three compared methods start from different small models:
\emph{depth} stands for a 3-layer model; \emph{length} stands for training with 128-token training data; \emph{compound} stands for a 6-layer model with 256-token training data.
}
\vspace{-0.4cm}
\label{fig:compound}
\end{figure*}
Further, we explore the potential choices of growth operators on each dimension.
We conduct controlled experiments and comprehensive analyses to compare various available solutions. 
These analyses further guide the design of effective compound growth operators. 
Specifically, we observe that, on the length dimension, embedding pooling is more effective than directly truncating sentences.
On the width dimension, parameter sharing outperforms low-rank approximation.

Guided by our analyses, we propose \method by combining the most effective growth operator on each dimension.
Experiments on standard benchmarks show that,
without sacrificing final performance, the final model speeds up the overall pre-training by 73.6\% and 82.2\% on BERT-base and BERT-large models respectively.

\begin{algorithm}[t]
\small
    \SetAlgoLined
    \caption{Progressive Training\label{alg:general}.
    \\$f$: network; $opt$: optimizer; $\gD$: dataset.
    \\$g_t$: the growth operator at stage $t$.
    \\$T$: the total number of growing stages.}
    $f_0 \gets opt(f_0, \xb, y)$ \\
    
    \For{$t \in [1, T]$}{

        $f_t \gets g_t(f_{t-1})$ %\tcp{\footnotesize\hspace{-0.2cm}\fontdimen2\font=1ex grow the model with operator $g_t$ at the $t$-th stage.}
        \For{$\xb, y \in \gD$} {
            $f_t \gets opt(f_t, \xb, y)$ %\tcp{\footnotesize\hspace{-0.2cm}\fontdimen2\font=1ex Parameter update}
        }
    }
    \textbf{return} Final network $f_T$
\end{algorithm}

% \begin{figure*}[t]
% \centering
%     \includegraphics[width=\textwidth]{img/diagram.pdf}
% \vspace{-0.3cm}
% \caption{Costs of different parts of a Transformer model. }
% \vspace{-0.4cm}
% \label{fig:diagram}
% \end{figure*}
% \input{src/relatedwork}
% \input{src/preliminaries}

\section{Progressive Compound Growth}
\noindent\textbf{Progressive Training.}
Algorithm~\ref{alg:general} presents a generic setup for progressive training. 
In each training stage $t$, the corresponding growth operator $g_t$ grows the model $f$. 
Then, $f$ is updated by the optimizer $opt$ before entering the next training step.
Correspondingly, our goal is to maximize the final model performance after all training stages, which can be formulated as minimizing the empirical loss $\gL$ over dataset $\gD$:  
\begin{align}
    \min_{g_t \in \gG}\; \gL(f_T)\;\; \st \;f_t = opt\left(g_t(f_{t-1}), \gD\right)
\label{eqn:obj}
\end{align}

\smallskip
\noindent\textbf{Compound Effect.}
Existing progressive training methods only focus on one model dimension. 
For example, \citet{Gong2019EfficientTO} conduct Transformer growth by gradually increasing the network \emph{depth}. 
\citet{devlin2019bert} use shorter input sequence \emph{length} at early stages.
However, as studies in network architecture search have revealed \citep{tan2019efficientnet}, growth operators that balance different model dimensions can achieve better performance than single-dimensional operators under the same budget.
Note that our objective (Equation~\ref{eqn:obj}) is close to the objective of EfficientNet~\citep{tan2019efficientnet}, which aims to find the optimal network architecture by maximizing the model accuracy for a given resource budget:
\begin{align*}
\label{eqn:efficient-net}
    \max_{d, w, r} \; Accuracy(\gN(d, w, r)) \;\; \\ \st  \;\;
    \mbox{\footnotesize \textit{Resource\_cost}}(\gN) \leq \mbox{\footnotesize target\_budget},
\end{align*}
where $\gN(d, w, r)$ is a CNN network, $d$, $w$, $r$ are coefficients to scale its depth, width, and resolution. 
In this work, 
we find that such a \emph{compound effect} also plays a vital role in progressive BERT training.
Intuitively, growing the network from more than one dimension creates larger potential to get better performance with less resource.
Restricting the growth operator from handling all dimensions would lead to inferior performance, as $\min_{g \in \gG}\; \gL(f_T) \geq \min_{g \in \gG \cup \gG^+}\; \gL(f_T)$.
% \end{remark}
The optimal value of the objective function (Equation~\ref{eqn:obj}) is bounded by the feasible set of the growth operator.

\smallskip
\noindent\textbf{Empirical Verification.}
For empirical verification, we compare existing single-dimensional growth operators in model depth and length with the corresponding compound operator that balances both dimensions.
For all three compared growth operators, their configurations are adjusted to make sure they have the same model after growth, and their low-cost models have empirically comparable training costs.
As to the training, we first train the low-cost model for 100/300/500/700K steps, and then grow the model to a standard BERT-base model for another 300K steps training.
For models trained with different steps/growth operators, we compare their performance after finetuning on MNLI, SQuaD v1.1, and SQuaD v2.0 respectively. 
As Figure \ref{fig:compound} shows, across different settings (columns) and metrics (rows), the compound operator consistently outperforms or at least achieves comparable results with single-dimensional operators.
The observation meets our intuition: to achieve same speedup, the compound method can distribute the reduction on training cost to different dimensions, and achieve better performance.

\section{Explore Possible Growth Operators}

\begin{table*}[t]
    \centering
    \caption{Empirical comparison among growth operators. For each operator, a low-cost model is first trained for 700K steps, then grown to the original BERT model for another 300K steps training.}
    \label{tb:control}
    \tightcaption{}
    \scalebox{0.78}{
    \renewcommand{\arraystretch}{.9}
      \begin{tabular}{l|ccccc|ccccc}
        \toprule
         & \multicolumn{5}{c|}{\textbf{BERT\textsubscript{base}}} & \multicolumn{5}{c}{\textbf{BERT\textsubscript{large}}} \\
         & \textbf{MNLI} & \multicolumn{2}{c}{\textbf{SQuAD v1.1}} &  \multicolumn{2}{c|}{\textbf{SQuAD v2.0}} & \textbf{MNLI} & \multicolumn{2}{c}{\textbf{SQuAD v1.1}} &  \multicolumn{2}{c}{\textbf{SQuAD v2.0}}  \\
         & \textbf{Acc.}  &  \ \ \textbf{EM} \ \  & \textbf{F1} & \ \  \textbf{EM} \ \  & \textbf{F1} & \textbf{Acc.}  &  \ \ \textbf{EM} \ \  & \textbf{F1} & \ \  \textbf{EM} \ \  & \textbf{F1} \\
        \midrule
        \textbf{Data Truncation}   & 83.72 & 82.72 & 90.00 & 76.06 & 79.18 & 85.80 & 85.51 & 92.18 & 79.56 & 82.57\\
        \textbf{Embed Pooling}      & 84.04 & 82.96 & 90.16 & 76.83 & 79.88 & 85.88 & 85.07 & 91.95 & 80.86 & 83.69\\
        \midrule
        \textbf{FFN Factorization} & 83.53 & 82.21 & 89.45 & 75.27 & 78.11 & 85.96 & 85.66 & 92.10 & 79.35 & 82.38\\
        \textbf{FFN Share Param.}       & 83.92 & 83.02 & 89.91 & 75.83 & 78.56 & 86.28 & 85.60 & 92.02 & 80.92 & 83.85\\
        \bottomrule        
    \end{tabular}%
    }
\end{table*}

% \begin{figure*}[t]
% \centering
%     \includegraphics[width=0.8\textwidth]{img/parameter_sharing.pdf}
% \vspace{-0.3cm}
% \caption{Demonstration of matrix factorization and parameter sharing on the width dimension. }
% \vspace{-0.4cm}
% \label{fig:parameter_sharing}
% \end{figure*}
After verifying the importance of compound growing, we conduct more analysis to provide guidance for growth operator design. 

% Since growth in the depth dimension has been thoroughly discussed in literature \citep{Gong2019EfficientTO,li2020shallow}, 
% here we focus our studies on other dimensions. 

\subsection{Length Dimension}
\noindent\textbf{Data Truncation} first limits the maximum length of input sequences by truncating the training sentences to a shorter length, and then train the model on full-length data.
Note that shorter input sequences usually come with less masked tokens to predict in each sentence.
For instance, \citet{devlin2019bert} first use sentences of at most 128 tokens (with 20 masked tokens) before training on data of 512 tokens (with 76 masked tokens).
The major issue of this data truncation operator is the incomplete update of position embeddings.
The model needs to learn embeddings for the extra positions from scratch at the last stage.

\noindent\textbf{Embedding Pooling.}
Inspired by the idea of multigrid training in the vision domain~\citep{wu2020multigrid}, we train the model with ``low-resolution text'' through embedding pooling over unmasked tokens.
Compared with data truncation, this method leaves the training data intact and can update all position embeddings. 
Specifically, since the output length of self-attention modules is decided by the length of query vectors, we only conduct pooling on query vectors in the first self-attention layer and keep key/value vectors intact.

% The existence of position embeddings gives Transformer the unique advantage in processing tokens regardless of their input orders,
% Thus, we first reorder the input token embeddings to separate masked tokens from unmasked ones in pre-training, and only apply pooling to the unmasked tokens.
% In this way, all masked tokens manage to preserve their unique representations for the masked language modeling task.
% Specifically, since the output length of self-attention modules is decided by the length of query vectors, we only conduct pooling on query vectors and keep key/value vectors intact.

As shown in the first group of Table \ref{tb:control}, data truncation (sequence length\tightmath{ = 256}) and mean pooling (\tightmath{k=2}) has similar performance on MNLI and SQuAD v1.1, while mean pooling outperforms data truncation on SQuAD v2.0.

\subsection{Width Dimension}
\label{subsec:width-ffn}
On the width dimension, we focus our study on the feedforward network module (FFN).
Similar to gradually increasing the network depth, one can also gradually increase the network width for Transformer growth. 
Specifically, the FFN module can be formed as $f(xW_1)W_2$, where $f(\cdot)$ is the activation function, $W_1 \in \mathbb{R}^{D \times H}$ and $W_2 \in \mathbb{R}^{H \times D}$ are parameters, $D$ and $H$ are the embedding size and the hidden size respectively.

\noindent\textbf{Matrix Factorization.}
A straightforward method is to approach the original weight matrix $W_i \in \mathbb{R}^{m \times n}$ by the product of two small matrices $W_{i1} \in \mathbb{R}^{m \times h}$ and $W_{i2} \in \mathbb{R}^{h \times n}$ in the early training stage.
In the late stage of training, we would recover $W_i$ as $W_{i1} \times W_{i2}$ and unleash the full potential.  

\noindent\textbf{Parameter Sharing.} 
Instead of decomposing original weight matrices with low-rank approximation, we try to employ parameter sharing by spliting the matrix into multiple blocks and sharing parameters across different blocks.
Formally, for input $x$,
\begin{equation}
\begin{array}{c}
{\scriptstyle
    f(x W_1)W_2 = f(x [W'_1, ..., W'_1]) \left[ \begin{array}{c} W'_2/k \\... \\ W'_2/k\end{array}\right]  = f(x W'_1)W'_2.
}
\end{array}
\end{equation}
Specifically, in the early training stage, we replace $W_1$ and $W_2$ with smaller matrices $W'_1 \in \mathbb{R}^{D \times \frac{H}{k}}$ and $W'_2 \in \mathbb{R}^{\frac{H}{k} \times D}$.
Then, at the growth step, we vertically duplicate (share) $W'_1$ for $k$ times along the dimension with size $H / k$ as the new $W_1$.
$W_2$ is generated similarly.
Similar to matrix factorization, this setting also preserves the output after the growth.
Random noise is added to $W_1$ and $W_2$ by the dropout layers in FFN, so that the shared small matrices will have different outputs and gradients in later training steps \citep{chen2015net2net}.

% Figure \ref{fig:parameter_sharing} demonstrates the difference between the two treatments.
As the second group of Table \ref{tb:control} shows, parameter sharing has significant superiority over matrix factorization with comparable budgets (k=4 for parameter sharing and h=0.2D for matrix factorization).

\subsection{Depth Dimension}
Transformer growth in the depth dimension has been thoroughly discussed in literature \citep{Gong2019EfficientTO,li2020shallow}.
Our observation in this dimension is consistent with their conclusions.
In experiments we also compare compound growth with the standard progressive stacking method.

\medskip
\noindent\textbf{Discussion.}
From the perspective of implementation, compound growth introduces little additional engineering effort compared with progressive stacking.
Specifically, the growth step of progressive stacking basically copies the parameters of the small model to corresponding layers of the full model.
The growth on the width dimension is a similar parameter copying process for the fully connected layers, while the growth on the length dimension removes the embedding pooling layer without changing any model parameters.

\begin{table*}[t]
    \centering
    \caption{The pre-training speedup and finetuning performance on dev sets of MNLI and SQuaD. 
    M/MM stands for matched/mismatched accuracy for MNLI. EM/F1 represents exact match score and F1 score for SQuaD.
    The FLOPs are estimated for forward pass operations, while the walltime is real training time profiled by the TensorFlow profiler from a distributed multi-host setting.}
    \label{tb:dev}      
    \tightcaption{}
    \scalebox{0.78}{
    \renewcommand{\arraystretch}{.8}
      \begin{tabular}{l | c | c | cc | cc | cc}
        \toprule
         & \textbf{speedup} & \textbf{speedup} & \multicolumn{2}{c|}{\textbf{MNLI Acc.}}   & \multicolumn{2}{c|}{\textbf{SQuAD v1.1}} &  \multicolumn{2}{c}{\textbf{SQuAD v2.0}}\\
         & \textbf{(FLOPs)} & \textbf{(walltime)} & \textbf{M} & \textbf{MM}  & \ \ \textbf{EM} \ \  & \textbf{F1} & \ \  \textbf{EM} \ \  & \textbf{F1}\\
        \midrule
        \textbf{BERT\textsubscript{BASE}} & -- & -- & 84.4 & 84.4 & 83.3 & 90.2 & 77.4 & 80.4  \\
        \textbf{Stack\textsubscript{BASE}} & +68.7\% & +64.9\% & 84.5 & 84.9  & 83.5 & 90.5 & 77.1 & 80.3\\
        \textbf{Compound\textsubscript{BASE}} & \textbf{+107.1\%} & +\textbf{73.6\%} & 84.7 & 84.7 & 83.8 & 90.3 & 77.0 & 80.0 \\
        \midrule
        \textbf{BERT\textsubscript{LARGE}} & -- & -- & 86.3 & 86.4 & 86.2 & 92.7 & 81.0 & 84.3\\
        \textbf{Stack\textsubscript{LARGE}} & +70.7\% & +69.7\% & 86.9 & 87.3 & 86.3 & 92.6 & 81.7 & 84.7\\
        \textbf{Compound\textsubscript{LARGE}} & \textbf{+111.4\%} & \textbf{+82.2\%} & 87.3 & 86.8 & 85.8 & 92.4 & 82.4 & 85.3\\
        \bottomrule        
    \end{tabular}%
    }
\end{table*}

\begin{table*}[t]
    \centering
    \caption{The test performance on the GLUE benchmark with metrics described in the original paper \citep{wang2018glue}, the higher the better. Compound stands for the proposed method with speedup shown in Table \ref{tb:dev}.}
    \label{tb:glue}
    \tightcaption{}
    \scalebox{0.8}{
    \renewcommand{\arraystretch}{.9}
      \begin{tabular}{l|ccccccccc|c}
        \toprule
         &  \textbf{\small CoLA} & \textbf{\small SST-2} & \textbf{\small MRPC} & \textbf{\small SST-B} & \textbf{\small QQP} & \textbf{\small MNLI-m/mm} & \textbf{\small QNLI}  & \textbf{\small RTE} & \textbf{\small WNLI} & \textbf{\small GLUE}  \\
        \midrule
        \textbf{\small BERT\textsubscript{BASE}} &52.1 & 93.5 & 88.9/84.8 & 87.1/85.8 & 71.2/89.2 & 84.6/83.4 & 90.5 & 66.4 & 65.1 & 78.3\\
        \textbf{\small Stack\textsubscript{BASE}} & 57.3 & 92.8 & 89.4/85.6 & 85.4/84.1 & 71.0/89.1 & 84.7/83.5 & 91.4 & 69.9 & 63.7 & 79.1\\
        \textbf{\small Compound\textsubscript{BASE}} & 50.1 & 92.6 & 89.1/85.2 & 85.4/83.9 & 70.9/88.9 & 84.6/83.6 & 91.3 & 70.1 & 65.1 & 78.3\\
        \midrule
        \textbf{\small BERT\textsubscript{LARGE}} & 60.5 & 94.9 & 89.3/85.4 & 87.6/86.5 & 72.1/89.3 & 86.7/85.9 & 92.7 & 70.1 & 65.1 & 80.5\\
        \textbf{\small Stack\textsubscript{LARGE}} & 62.2 & 94.3 & 89.9/85.9 & 86.0/85.0 & 71.2/88.9 & 86.9/86.3 & 93.0 & 75.2 & 65.1 & 81.1\\
        \textbf{\small Compound\textsubscript{LARGE}} & 61.2 & 94.2 & 90.2/86.7 & 86.4/85.7 & 71.4/89.2 & 87.2/86.1 & 93.6 & 73.3 & 65.8 & 81.1\\
        \bottomrule        
    \end{tabular}%
    }
\end{table*}

\section{Experiment}

\noindent\textbf{Experiment Setups.} 
We train the original BERT models following the same settings in \cite{devlin2019bert} with 256 batch size and 512-token data.
All compared models will finally grow to the original model, and keep the total number of training steps to 1M.
We evaluate the final model on the GLUE benchmark~\citep{wang2018glue} including 9 subtasks, and the two versions of SQuAD~\citep{rajpurkar2018know} datasets for question answering. 
More detailed experiment settings can be found in the appendix for reproduction.

\noindent\textbf{Compared Methods.} 
Previous studies have rarely focused on progressive Transformer growth for BERT training, and progressive Transformer stacking~\citep{Gong2019EfficientTO} is the only directly comparable method to the best of our knowledge.
We apply their method on the official BERT model with the same training setting, learning rate schedule and hardware as our method.
We set the training schedule as 300K steps with \sfrac{1}{4} number of layers, 400K steps with \sfrac{1}{2} number of layers, and 300K steps with the full model.

\noindent\textbf{Our Method.} For \method{}, we apply treatments on three dimensions for the low-cost model:
(1) mean embedding pooling with size 2 on the length dimension; 
(2) parameter sharing with $k=2$ on FFN modules on the width dimension; 
(3) stacking on the depth dimension.
Following the same setting as compared methods, we try to equally distribute the 1M training steps.
We train the model with all treatments with \sfrac{1}{4} number of layers and \sfrac{1}{2} number of layers for 200K steps respectively, and then stack it to full layers with treatments on the width and length dimensions for another 300K steps.
At the last stage, we train the full model for 300K steps, just like the compared method.

\noindent\textbf{Results.}
Table \ref{tb:dev} shows the speedup of different models.
We estimate the inference FLOPs for compared models and get their real training time from the Tensorflow profiler
\footnote{https://www.tensorflow.org/guide/profiler}. 
On the BERT-base model, stacking and \method{} speeds up pre-training by $68.7\%$ and $107.1\%$ respectively in FLOPs, $64.9\%$ and $73.6\%$ respectively on walltime.
On the BERT-large model, stacking and \method{} speeds up pre-training by $70.7\%$ and $111.4\%$ respectively in FLOPs, $69.7\%$ and $82.2\%$ respectively on walltime.
Though \method{} is significantly faster, on development sets of MNLI and SQuaD, the compared methods do not have significantly different finetuning performance from the original BERT models.

Table \ref{tb:glue} shows the test performance on the GLUE benchmark.
Both compared methods achieve at least the same performance as the original BERT model.
While \method{} saves more training time, it achieves the same performance with stacking on the large model.
On the base model, stacking is better in terms of average GLUE score, mainly due to its advantage on the CoLA dataset.
Such an unusual gap on CoLA might be caused by its relatively small volume and corresponding random variance~\citep{dodge2020fine}.
On the larger and more robust MNLI dataset, the compared methods achieve almost the same score.

\section{Related Work}
\label{sec:related}

Progressive training was originally proposed to improve training stability, which starts from an efficient and small model and gradually increase the model capacity~\citep{simonyan2014very}. 
Recent study leverages this paradigm to accelerate model training. 
For example, multi-level residual network~\citep{Chang2017MultilevelRN} explores the possibility of augmenting network depth in a dynamic system of view and transforms each layer into two subsequent layers.
AutoGrow~\citep{wen2020autogrow} attempts to automate the discover of proper depth to achieve near-optimal performance on different datasets. 
LipGrow~\citep{dong_lipgrow_2020} proposes a learning algorithm with an automatic growing scheduler for convolution nets.
At the same time, many studies have been conducted on the model growing operators. 
Network Morphism~\citep{wei2016network, wei2017modularized} manages to grow a layer to multiple layers with the represented function intact. 
Net2net~\citep{chen2015net2net} is a successful application to transfer knowledge to a wider network with function-preserving initialization.
Similar ideas can be discovered in many network architectures, including progressive growing of GAN~\citep{karras2017progressive} and Adaptive Computation Time~\citep{graves2016adaptive, jernite2016variable}.

As large-scale pre-training keeps advancing the state-of-the-art~\citep{devlin2019bert,Radford2018ImprovingLU}, their overwhelming computational consumption becomes the major burden towards further developing more powerful models~\citep{Brown2020LanguageMA}.
Preliminary application of progressive training has been made on Transformer pre-training.
\cite{devlin2019bert} designs two-stage training with a reduced sequence length for the first 90\% of updates. 
\cite{Gong2019EfficientTO} stack shallow model trained weights to initialize a deeper model, which grows the BERT-base model on the depth dimension and achieves 25\% shorter training time. 

\section{Conclusion}
In this work we empirically verify the importance of balancing different dimensions in Transformer growth and propose compound growth operators, which
integrates operators for more than one dimension.
Moreover, we conduct controlled experiments on various design choices of growth operators, which provides a practical guidance to algorithm design.
Our final model speeds up the training of the BERT-base and BERT-large models by $73.6\%$ and $82.2\%$ in walltime respectively while achieving comparable performance.

% \section*{Acknowledgements}

% Entries for the entire Anthology, followed by custom entries
\bibliography{naacl2021}
\bibliographystyle{naacl2021}

\newpage
\appendix

\newpage\newpage

\section{Experiment Details}
All our models are implemented based on the TensorFlow implementation\footnote{\url{https://github.com/tensorflow/models/blob/master/official/nlp/modeling/models/bert\_pretrainer.py}} of BERT~\citep{tensorflowmodelgarden2020} and trained on TPU v3 with 64 chips. 
We keep the original WordPieceTokenizer and original position embeddings (instead of relative position encoding used in~\cite{dai2020funnel}).
Following \cite{devlin2019bert}, we use the English Wikipedia corpus and the BookCorpus for pre-training.
For each finetuning task, we search hyperparameters from following candidates: batch size=16/32/64, learning rate=3e-4/1e-4/5e-5/3e-5.

\noindent\textbf{Optimization.}
The original BERT models use the AdamW~\citep{loshchilov2018decoupled} optimizer with learning rate decay from 0.0001 to 0 and 10K steps of warmup~\cite{Liu2020OnTV}.
At the start of each progressive training stage, the learning rate is reset to 0.0001 and keeps decaying as the original schedule.

\noindent\textbf{Baseline Implementation.}
We apply the compared stacking method \citep{Gong2019EfficientTO} on the official BERT model with the same training setting, learning rate schedule and hardware as our method, and achieves better performance than the reported numbers in the original paper.
To further unleash the potential of the compared method, we adjust their original training schedule to 300K steps with \sfrac{1}{4} number of layers, 400K steps with \sfrac{1}{2} number of layers, and 300K steps with the full model.
The new training schedule is much faster than the reported one (speedup from the reported +25\% to +64.9\%) and still gives better final performance than the original paper.
This is the fastest stacking model we can get without performance drop.

\begin{figure}[t]
\centering
    \includegraphics[width=0.35\textwidth]{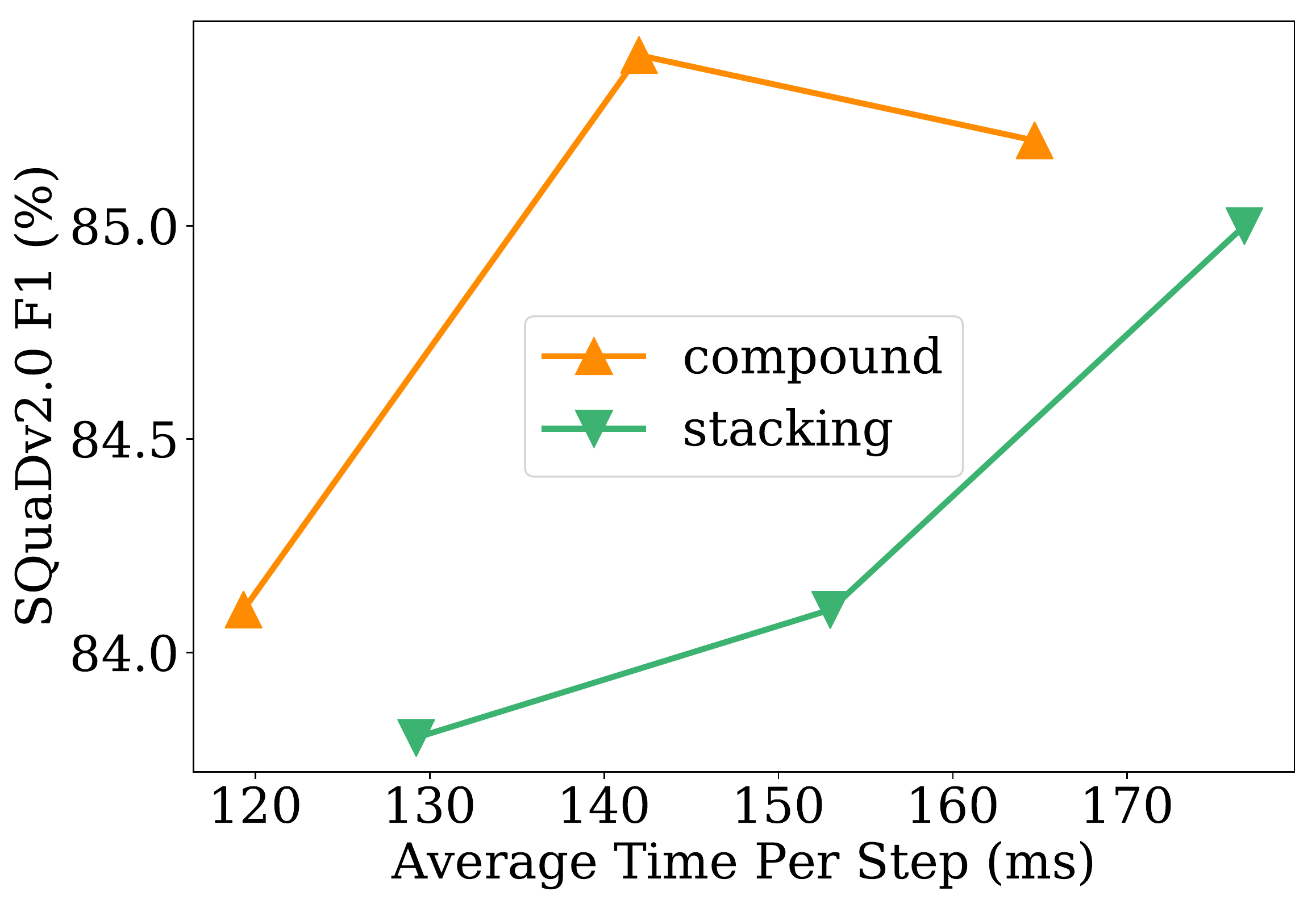}
% \vspace{-0.3cm}
\caption{Compare the speed-performance trade-off of stacking and \method{} on BERT\textsubscript{large}. The three data points in each curve is generated with 300K/500K/700K low-cost training steps, respectively.}
\vspace{-0.4cm}
\label{fig:tradeoff}
\end{figure}
\section{Further Comparison Between \method{} and Stacking}
To have a deeper understanding of the compared methods, we study their speed-performance trade-off by adjusting the training schedule.
Specifically, each time we reduce 200K low-cost training steps for both models, and compare their validation F1 score on SQuaDv2.0.
As Figure \ref{fig:tradeoff} shows, \method{} has clear performance advantage when given comparable training budgets, which further verifies our hypothesis.

\end{document}